\documentclass[runningheads]{llncs}
\usepackage[utf8]{inputenc}
\usepackage[T1]{fontenc}
\usepackage{graphicx}
\usepackage{amssymb}
\usepackage{color}
\usepackage{diagbox}

\begin{document}
\title{Colonoscopy Coverage Revisited: Identifying Scanning Gaps in Real-Time}
%
%
\author{
G. Leifman \and
I. Kligvasser \and
R. Goldenberg \and
M. Elad \and
E. Rivlin}

\authorrunning{G. Leifman et al.}
%
\institute{Verily}

\maketitle              
%

%
%
\begin{abstract}
Colonoscopy is the most widely used medical technique for preventing Colorectal Cancer, by detecting and removing polyps before they become malignant. 
Recent studies show that around 25\% of the existing polyps are routinely missed. 
While some of these do appear in the endoscopist’s field of view, others are missed due to a partial coverage of the colon. 
The task of detecting and marking unseen regions of the colon has been addressed in recent work, where the common approach is based on dense 3D reconstruction, which proves to be challenging due to lack of 3D ground truth and periods with poor visual content. 
In this paper we propose a novel and complementary method to detect deficient local coverage in real-time for video segments where a reliable 3D reconstruction is impossible.
Our method aims to identify skips along the colon caused by a drifted position of the endoscope during poor visibility time intervals. 
The proposed solution consists of two phases. 
During the first, time segments with good visibility of the colon and \textit{gaps} between them are identified. 
During the second phase, a trained model operates on each \textit{gap}, answering the question: "Do you observe the same scene before and after the gap?" 
If the answer is negative, the endoscopist is alerted and can be directed to the appropriate area in real-time. 
The second phase model is trained using a contrastive loss based on an auto-generated examples.
Our method evaluation on a dataset of 250 procedures annotated by trained physicians provides sensitivity of 75\% with specificity of 90\%.

\end{abstract}

\section{Introduction}

Colorectal cancer is one of the most preventable cancers, as early detection and through screening is highly effective. 
The most common screening procedure is optical colonoscopy -- visually examining the surface of the colon for abnormalities such as colorectal lesions and polyps. 
However, performing a thorough examination of the entire colon surface is proven to be quite challenging due to unavoidable poor visibility segments of the procedure. As a consequence, improperly inspected regions may lead to a lower detection rate of polyps. Indeed, recent studies have shown that approximately 25\% of polyps are routinely missed during a typical colonoscopy procedure~\cite{kim2017miss}.

Various efforts to automatically detect and mark non-inspected regions of the colon are reported in recent publications, where the common approach relies on the creation of a dense 3D reconstruction of the colon's shape~\cite{chen2019slam,freedman2020detecting,posner2022c,rau2019implicit,shao2022self,zhang2020template}. 
However, such a reconstruction based on video solely is a challenging task, and especially so in colonoscopy, in which reflections, low-texture content, frequent changes in lighting conditions and erratic motion are common. As a consequence, while the above 3D approach has promise, it is limited to segments of the video exhibiting good visual quality.

\begin{figure}
    \begin{center}
    \includegraphics[width=3.5in]{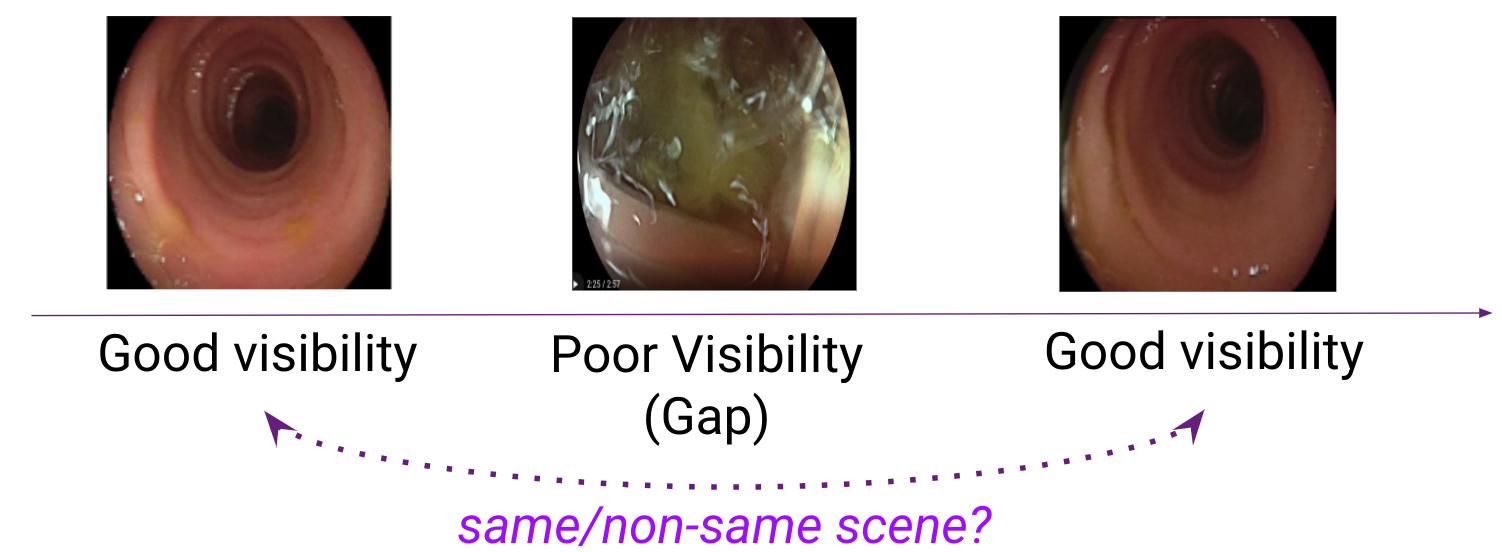}
    \end{center}
    \caption{
    Our solution starts by detecting time segments with good visibility of the colon and gaps between them. For each such gap we answer the question: \emph{Do you observe different scenes before and after the gap?} If the answer is positive, the endoscopist is alerted to revisit the area in real-time.
    } 
    \label{fig1}
\end{figure}

In this work we propose a novel real-time approach for detecting deficient local coverage, complementing the 3D reconstruction methods mentioned above. Our proposed strategy 
provides a reliable, stable and robust solution for the grand challenge posed by temporal periods of poor visual content, such as camera blur, poor camera positioning, occlusions due to dirt and spayed water, and more. The proposed method consists of two main phases. During the first, we identify time segments with good visibility of the colon and gaps of poor visibility between them. For this purpose we train a binary classifier, leveraging a small set of annotated images and a self-supervised training scheme. 
During the second phase, we train an ML model that aims to answer the following question for each gap: \emph{Do you observe different scenes before and after the gap?} (see Figure~\ref{fig1}). 
If the answer is positive, we suspect a loss of coverage due to an unintentional drift of the endoscope position, and therefore alert the endoscopist accordingly in real-time to revisit the area. 

The second phase model is designed to generate low-dimensional frame-based descriptors that are used for scene-change detection via a simple Cosine distance evaluation. 
This network is trained using a contrastive loss based on automatically generated positive and negative pairs of video segments.
These training examples are sampled from good-visibility segments of real colonoscopy videos, where 
the translational speed of the endoscope can be reliably estimated.

To evaluate our method we introduce a dataset of 250 colonoscopy procedures (videos).
Two doctors have been asked to evaluate up to 5 gaps per video and decide whether they suspect loss of coverage there. 
The evaluation of our method using this annotated dataset provides sensitivity of 75\% with specificity of 90\%.

We note that our task of same-scene detection in the colon is related to image retrieval~\cite{image_retrival_ali2019efficient,image_retrival_radenovic2018fine,image_retrival_yan2020deep} and geo-localization~\cite{geo_berton2021viewpoint,geo_lin2013cross,geo_lin2015learning}.
There is also some similarity to techniques employed for face recognition~\cite{face_adjabi2020past,face_bae2023digiface,face_deng2019arcface,face_kortli2020face} and person re-identification~\cite{re_id_chen2021joint,re_id_lin2020unsupervised,re_id_wang2020unsupervised}.
In the narrower domain of colonoscopy, the only closely related work we are aware of is reported in~\cite{ma2021colon10k}. 
While their technique for location recognition is related to our scene descriptor generation, their eventual tasks are markedly different, and so are the evaluation protocols. Nevertheless, for completeness of this work, we evaluate our scene descriptors on their dataset and show that our method outperforms their results.

To summarize, this work offers three main contributions:
\begin{itemize}
    \item We present a novel stable, robust and accurate method for detecting deficient local coverage in real-time for periods with poor visual content.  

    \item Our coverage solution complements the 3D reconstruction approach, covering cases beyond it's reach; 
    
    \item We introduce a novel self-supervised method for generating frame-based descriptors for scene change-detection in colonoscopy videos. 
    
\end{itemize}

\noindent This paper is organized as follow:
Section~\ref{Good_Visibility_Section} describes Phase I of our method, aiming to identify time segments with good visibility of the colon and gaps between them. 
Phase II of our method is presented in Section~\ref{Gaps_with_Loss_of_Coverage}, addressing the \emph{same-scene} question by metric learning. 
Section~\ref{Results} summarizes the results of our experiments and Section~\ref{Conclusion} concludes the paper.

\section{Method: Phase I -- Visibility 
Classification}
\label{Good_Visibility_Section}

Our starting point is a frame-based classification of the visibility content. We characterize  good visibility frames as those having a clear view of the tubular structure of the colon. In contrast, poor visibility frames may include severe occlusions due to dirt or sprayed water, a poor positioning of the camera - being dragged on the colon walls, or simply blurred content due to rapid motion. 

\begin{figure}
    \centering
    \includegraphics[width=0.8\textwidth]{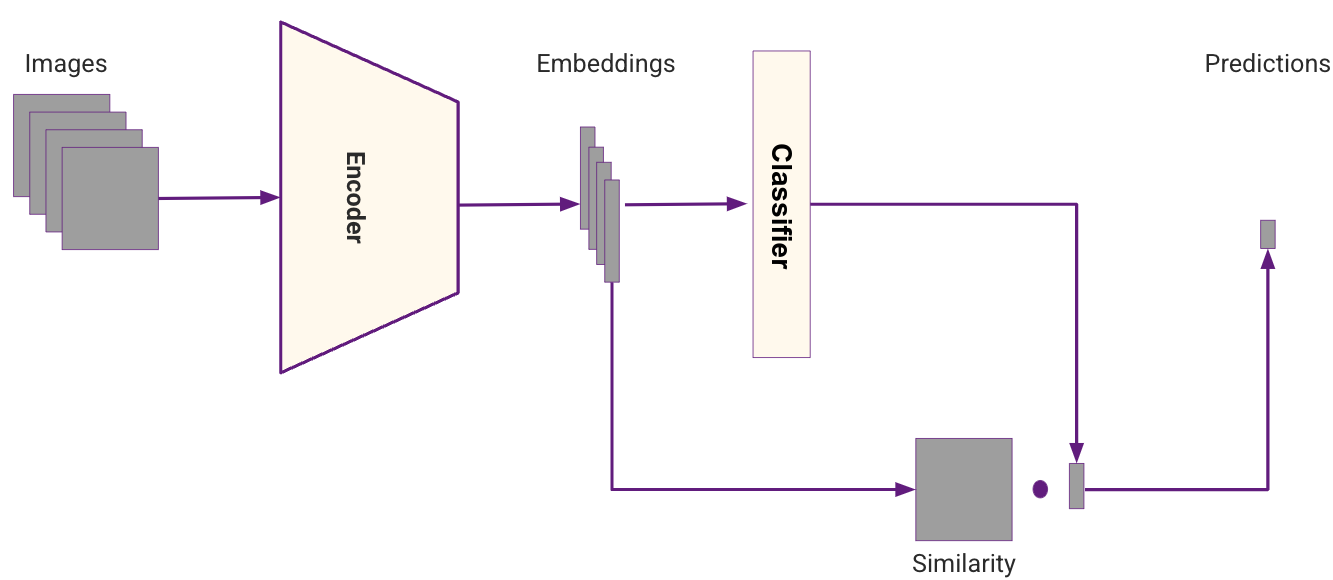}
    \caption{
    To achieve high accuracy visibility classifier, we train an encoder in an unsupervised manner and then train a binary classifier the resulting embeddings using the labeled data. 
    Further improvement is made by smoothing predictions based on similarity distances, resulting in $94\%$ accuracy on the validation set.
    } 
    \label{fig_visibility_classifier}
\end{figure}

In order to solve this classification task, we gather training and validation annotated datasets by experts. Operating on 85 different colonoscopy videos, 5 good visibility segments and 5 poor ones were identified in each. A naive supervised learning of a classifier leads to an unsatisfactory $84\%$ accuracy on the validation set due to insufficient data. In an attempt to improve this result, we adopt a semi-supervised approach. First, we pre-trained an encoder on large (1e6) randomly sampled frames using simCLR~\cite{chen2020simple}. This unsupervised learning embeds the frames such that similar ones (obtained by augmentations of the same frame) are close-by, while different frames (the rest of the frames in the batch) are pushed away. Given the learned encoder, we train a binary classifier on the resulting embeddings using the labeled data. Since the dimension of the embedding vectors is much smaller then the original frame sizes ($512$ vs. $224^2$), this approach leads to far better accuracy of $93\%$.
We further improve the above  
by smoothing the predictions based on their embeddings, as shown in Fig.~\ref{fig_visibility_classifier}. 
For each input batch of 512 frames, their cross-similarities (the cosine distance between their embedding vectors) are leveraged, such that similar frames are also encouraged to be assigned to the same class. 
This improves the per-frame accuracy on the validation set up to $94\%$. 

To conclude, the trained classifier provides a partitioning of the time axis into disjoint intervals of good or poor visibility.
In order to further relax these intervals, we apply a median filter with window size of 10 frames.


\section{Method: Phase II -- Gaps with Loss of Coverage}
\label{Gaps_with_Loss_of_Coverage}

After partitioning the procedure timeline into periods with good visibility and gaps between them, our goal now is to identify gaps with a potential loss of coverage, defined as exhibiting a change of the scene between their ends. In order to compare scenes before and after a gap, we learn distinctive frame descriptors. These vectors are compared via a simple distance measure for addressing the same/not-same scene question. 
While the direct approach towards this task would be to gather a training set of many thousands of such gaps along with their human annotation, we introduce a much cheaper, faster, and easier alternative based on a self-supervised approach. In this section we describe all these ingredients in greater details.



\noindent {\bf Scene Descriptors:} Assume that a training set of the form $\{F_1^k,F_2^k,c_k\}_{k=1}^N$ is given to us, where 
$F_1^k$ and $F_2^k$ are two frames on both sides of a given gap, and $c_k$ is their label, being $c_k=1$ for the same scene and $0$ otherwise. $N$ is the size of this training data, set in this work to be $N=1e5$ examples. We design a neural network $f=T_\Theta (F)$ that embeds the frame $F$ to the low-dimensional vector $f\in \mathbb{R}^{512}$, while accommodating our desire to serve the same/not-same scene task. More specifically, our goal is to push same-scene descriptor-pairs to be close-by while forcing pairs of different scenes to be distant, being the essence of  contrastive learning, which has been drawing increased attention recently~\cite{bachman2019learning,chen2020simple,he2020momentum,wang2021understanding}. Therefore, we train $T_\Theta (\cdot)$ to minimize the loss function
\begin{eqnarray}
L(\Theta) & = & \sum_{k=1}^N (2c_k-1) d\left(T_\Theta (F_1^k),T_\Theta(F_2^k\right) \\
& = & \sum_{\{c_k=1\}_k}  d\left(T_\Theta (F_1^k),T_\Theta(F_2^k\right) - \sum_{\{c_k=0\}_k}  d\left(T_\Theta (F_1^k),T_\Theta(F_2^k\right) \nonumber.
\end{eqnarray}
In the above expression, $d(\cdot,\cdot)$ stands for a distance measure. In this work we use the Cosine similarity $d(f_1,f_2) = 1-f_1^T f_2/\|f_1\|_2\|f_2\|_2$. 


\begin{figure}[b]
    \centering
    \includegraphics[width=0.8\textwidth]{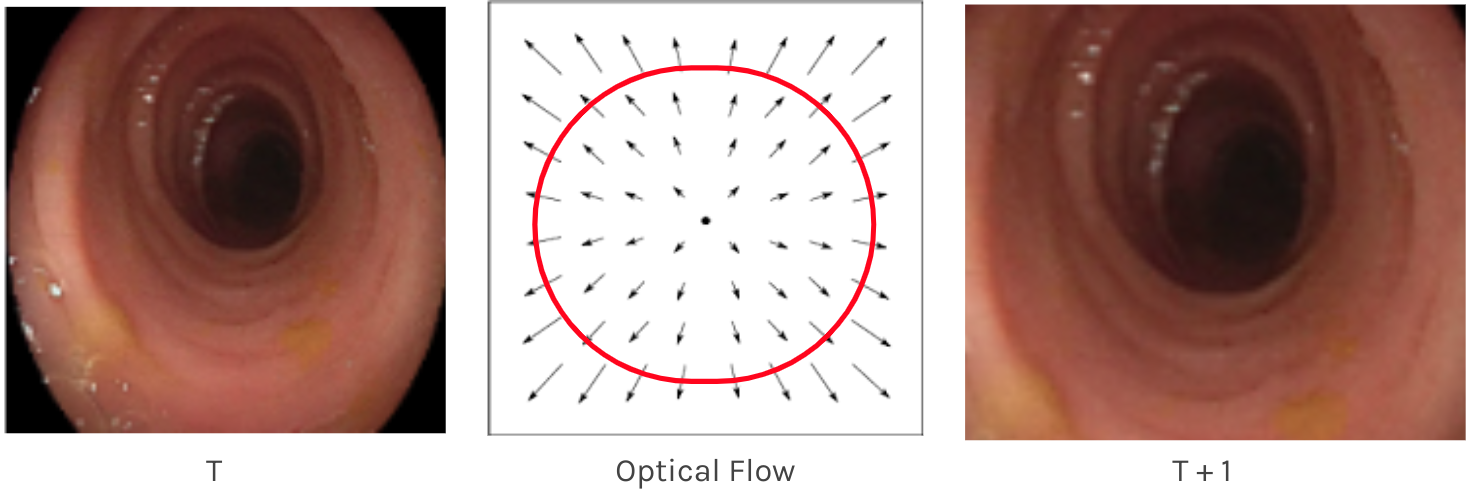}
    \caption{
    Endoscope displacement estimation is based on optical-flow calculation between consecutive frames using the amount of flow trough the frame boundary (see~\cite{kelner2022motion}). 
    }
    \label{fig:optical_flow}
\end{figure}

\noindent {\bf Creating the Training Data:} 
Constructing the training set $\{F_1^k,F_2^k,c_k\}_{k=1}^N$ might be a daunting challenge if annotations by experts are to be practiced. We introduce a fully automatic alternative that builds on a reliable displacement estimation of the endoscope, accessible in good visibility video segments of any real colonoscopy. 
This displacement can be evaluated by estimating the optical-flow between consecutive frames (see~\cite{oliveira2021registration,sun2018pwc}) and estimating the amount of flow trough the frame boundary~\cite{kelner2022motion} (see Figure~\ref{fig:optical_flow}). 

\begin{figure}
    \includegraphics[width=\textwidth]{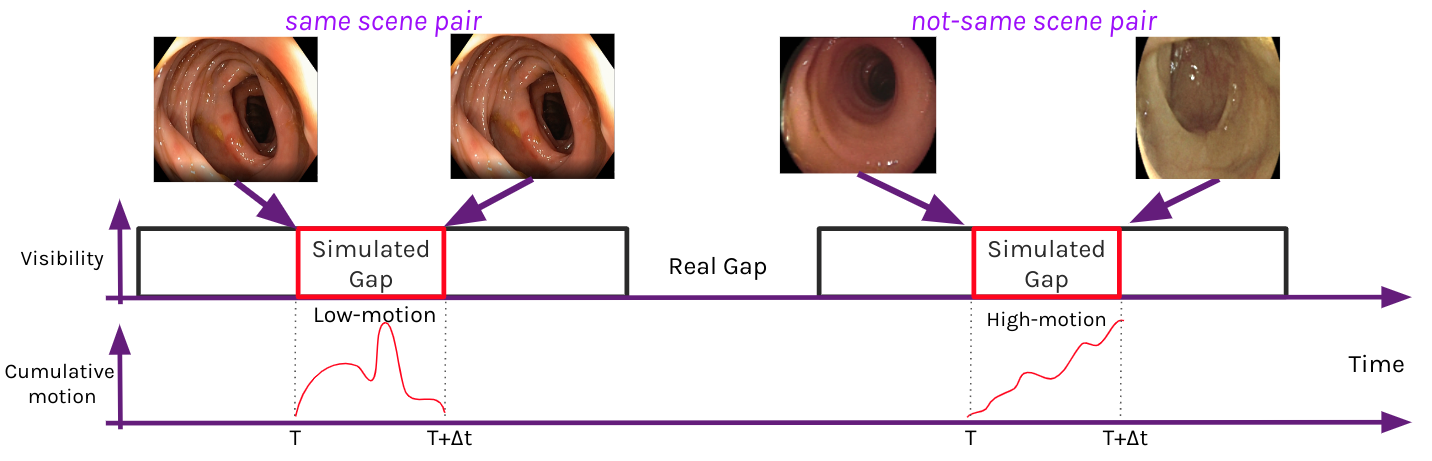}
    \caption{
    We simulate random artificial gaps of various duration in good-visibility video segments, estimate the endoscope motion within these simulated gaps, 
    and get this way reliable training examples for our overall task.
    Gaps associated with low accumulated motion contribute a `same-scene' training example ($c_k=1$), while high-motion gaps refer to a different scene pair ($c_k=0$).
    }
    \label{fig:cumulative_motion_pseudo_gaps}
\end{figure}

Given any time interval of good visibility content, the cumulative directional transnational motion can be estimated rather accurately. 
Thus, starting with such a video segments, and randomly marking an inner part of it of a random length of $5-30$ seconds as a pseudo-gap, we can define frames on both its ends as having the same scene or not based on the accumulated displacement. 
Figure~\ref{fig:cumulative_motion_pseudo_gaps} presents the whole process of creating training examples this way, easily obtaining triplets $\{F_1^k,F_2^k,c_k\}$.

Our attempts to improve the above contrastive training scheme by introducing a margin, as practiced in~\cite{hadsell2006dimensionality} and employing a ``soft-max'' loss version~\cite{wang2021understanding}, did not bring a significant improvement.
A technique that delivered a benefit is to pre-train the network $T_\Theta$ in fully unsupervised way using simCLR~\cite{chen2020simple} (as in Section~\ref{Good_Visibility_Section}), and proceed along the above contrastive learning scheme.
Implementation details can be found in the supplementary material.

\noindent {\bf Gap Classification:} With a simple machinery of a distance evaluation of the frame descriptors on both ends of any gap, we are now equipped to answer our main questions: \textit{Is there a potential loss of coverage during this poor-visibility video segment? Has the probe drifted away form its original position?} As this distance evaluation can be applied over various frames on both sides of the gap, the challenge is to find a reliable fusion of the information within these many pairs of frames. While we have experimented with various such options, 
the best results are achieved by calculating a single descriptor for the scenes before and after the gap, and then comparing these using a Cosine distance. This unified descriptor, $\bar{f}$, is obtained by a weighted average of the individual descriptors in a segment of $2$ seconds on each side, $f_i$, as follows:
$\bar{f} = \sum_i{f_i w_i} / \sum_i{w_i} $, 
where  $w_i = v_i e^{-s_i}$, $v_i$ and $s_i$ are the raw visibility score and the temporal distance to the gap, both referring to the $i$-th frame. 
While the effectiveness of employing such a simple averaging of the descriptors might seem surprising, a similar strategy was proven successful for face recognition from multiple views in~\cite{wolf2008descriptor}.


\section{Results}
\label{Results}
As explained in Section~\ref{Gaps_with_Loss_of_Coverage}, first we generate per-frame scene descriptors and then employ them to detect the gaps with potential loss of coverage.
This section starts from presenting the evaluation of the stand-alone scene descriptors and compares them to SOTA.
Then we describe the dataset of the annotated gaps and present the evaluation of our gap classifier on this dataset.

\noindent {\bf Scene Descriptors:}
We evaluate our scene descriptors on the recently released dataset for colonoscopic image retrieval -- Colon10K~\cite{ma2021colon10k}.
This dataset contains 20 short sequences (10,126 images), where the positive matching images were manually labeled and verified by an endoscopist.
We follow the setup and the evaluation metrics described in~\cite{ma2021colon10k}.
In total, they have 620 retrieval tasks (denoted by ``\textit{all}''), while 309 tasks use the intervals that are not direct neighbor frames of their queries as positives (denoted by ``\textit{indirect}'').
We use the data from Colon10K for the evaluation purposes only.
Table~\ref{table:colon10k_results} compares the results to those reported in~\cite{ma2021colon10k}.
\textit{Rank-1 recognition rate} is the percentage of tasks in which the most similar to the query image is true positive. The
\textit{Mean average precision} is the area under the precision-recall curve.
For both metrics our method outperforms~\cite{ma2021colon10k} for both ``\textit{all}'' and ``\textit{indirect}'' tasks.

\begin{table}[t]
\centering
\begin{tabular}{ |p{1.2cm}|p{2.2cm}|p{2.2cm}|p{2.2cm}|p{2.2cm}| } 
 \hline
 & \multicolumn{2}{|c|}{Rank-1 recognition rate} & \multicolumn{2}{|c|}{Mean average precision (mAP)} \\
 \hline
 Method& \textit{all} & \textit{indirect}  & \textit{all} & \textit{indirect} \\
 \hline
 ~\cite{ma2021colon10k} & 0.9032 & 0.8058 & 0.9042 & 0.8245 \\
 Our & 0.9131 & 0.8173  & 0.9723 & 0.9112 \\
 \hline
\end{tabular}
\caption{Comparison of our scene descriptor generation to~\cite{ma2021colon10k} on the Colon10K dataset. In all the evaluated metrics our method outperforms~\cite{ma2021colon10k}.}
\label{table:colon10k_results}
\end{table}

\begin{figure}[b]
    \centering
    \includegraphics[width=0.8\textwidth]{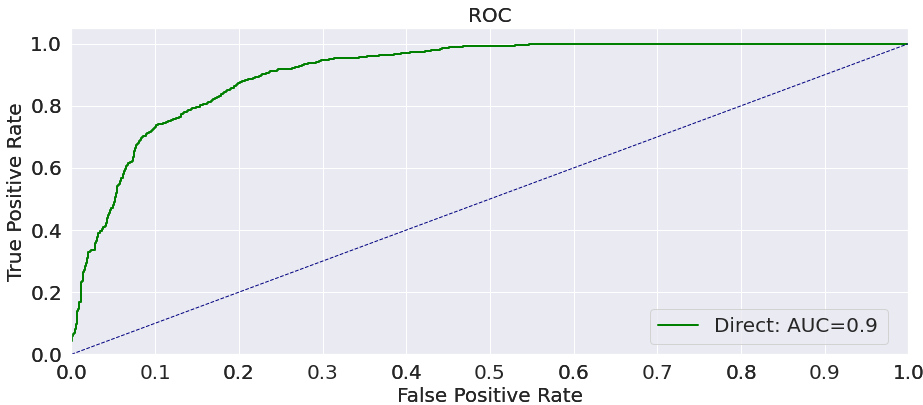} 
    \caption{Direct gap classification: ROC curve evaluated on the whole dataset (750 gaps).} 
    \label{fig:roc}
\end{figure}

\noindent {\bf Gap Classification:}
\label{Gap_Classification_results}
In order to evaluate our gap classification we introduce a dataset of 250 colonoscopy procedures (videos) from five different hospitals.
We have automatically identified between 2 to 5 \emph{true} gaps in each video and presented these to doctors for their annotation --  whether a loss of coverage is suspected. 
Each gap was evaluated by two doctors and the ones without a consensus ($\sim$25\%) were omitted. This resulted with 750 gaps having high-confidence annotations, 150 of which are marked as exhibiting a loss of coverage.
Figure~\ref{fig:roc} presents the ROC of our direct gap classification method evaluated on the whole dataset of 750 gaps.
At the working point of 10\% false alarms (alert on gap with no coverage loss) we cover $75$\% of gaps with real coverage loss.
The area under curve (AUC) is 0.9, which  usually indicates a high-accuracy classifier.

The above classification exploits the information before and after the gap, while completely disregarding the information about the gap itself. Having a dataset of annotated true gaps, we can improve this accuracy by a \textit{supervised} learning that exploits the gap characteristics.
We thus split the dataset of the annotated gaps 50:50 to training and evaluation.
Since we have a very limited amount of the training examples we use a low-dimensional classifier -- Gradient Boosting~\cite{brownlee2016xgboost} -- that operates on the following input data: (i)  A 32-bin histogram of the similarity matrix' values between frames 2 seconds before and after the gap; (ii) A 32-bin histogram of the visibility scores 2 seconds before and after the gap; (iii) A 32-bin histogram of the visibility scores inside the gap; and (iv) The duration of the gap. 
We performed class-balancing using up-sampling with augmentations before training.

Table~\ref{table:comparison_direct_supervised} compares the original approach to the supervised one, summarising the contribution of different input features to the final accuracy (measured by AUC).
In the supervised approach we use one half of the dataset for training, thus the evaluation is performed using the other half of the dataset for both the original and the supervised approaches. 
In the first approach we also explored a classificaiton based on the gap duration only, getting an AUC of 0.651, being higher than random but lower than employing frame similarities.
Weighing the scene descriptors by the visibility scores (see Section~\ref{Gaps_with_Loss_of_Coverage}) improves the AUC by 2\%.
In the supervised approach both gap duration and visibility scores inside the gap provide a substantial contribution of 2\% each to the AUC.

\begin{table}[t]
\centering
\begin{tabular}{ |c|c|c|c|c|c| } 
\hline

\backslashbox{Method}{Features}    & Frame       & Gap      & Visibility     & Visibility & AUC \\
    & similarities &  duration & inside the gap & outside the gap &\\
\hline
                    &  & \checkmark & & & 0.651 \\
  Original   & \checkmark &  & & & 0.876 \\
                  & \checkmark &  & & \checkmark & 0.896 \\
\hline
   & \checkmark & & & & 0.881\\
  Supervised & \checkmark& \checkmark& & & 0.898\\
   & \checkmark& \checkmark& \checkmark& & 0.929 \\
   & \checkmark& \checkmark& \checkmark& \checkmark& 0.932\\
 \hline
\end{tabular}
\caption{Impact of various features on the AUC, evaluated on 375 test gaps.} 
\label{table:comparison_direct_supervised}
\end{table}

\section{Conclusion}
\label{Conclusion}
This work presents a novel method for the detection of deficient local coverage in real-time for periods with poor visual content, complementing any 3D reconstruction alternative for coverage assessment of the colon. 
Our method starts with an identification of time segments with good visibility of the colon and \textit{gaps} between them. 
For each such \textit{gap} we train an ML model that tests whether the scene has changed during the gap, alerting the endoscopist in such cases to revisit a given area in real-time. Our learning constructs frame-based descriptors for same scene detection, leveraging a self-supervised approach for generating the required training set. For the evaluation of the gap classification results we have built a dataset of 250 colonoscopy videos with annotations of gaps with deficient local coverage. 
Our future work includes an extension of our approach to a guidance of the endoscopist to the exact place where the coverage was lost, and using our scene descriptors for bookmarking points of interest in the colon. 

%
%
%
%
\bibliographystyle{splncs04}
\bibliography{main}
\end{document}